\documentstyle[bigpage,epsfig]{article}
\author{Caroline Lyon\\
Computer Science Department\\ University of Hertfordshire\\
Hatfield AL10 9AB, UK\\
{\tt C.M.Lyon@herts.ac.uk}
\and
Stephen Brown\\
Mathematics Department\\ University of Hertfordshire\\
Hatfield AL10 9AB, UK\\
{\tt S.Brown@herts.ac.uk}
}
\date{}
\title{ Evaluating Parsing Schemes with Entropy Indicators}
\begin{document}
\maketitle
\begin{abstract}
This paper introduces an objective metric for evaluating
a parsing scheme. It is based on Shannon's original work with letter
sequences, which can be extended to part-of-speech tag sequences. It is shown
that this regular language is an inadequate model for natural language,
but a representation is used that models language slightly higher in the
Chomsky hierarchy. 

We show how the entropy of parsed and unparsed sentences can be measured.  
If the entropy of the parsed sentence is lower, this indicates that
some of the structure of the language has been captured. 

We apply this entropy indicator to support one particular parsing
scheme that effects a top down segmentation. This approach could 
be used to decompose the parsing task into computationally more tractable 
subtasks. It also lends itself to the extraction of predicate/argument
structure.
\end{abstract}
\section{Introduction}
This paper introduces an objective metric for assessing the effectiveness
of a parsing scheme. Information theoretic indicators can be used to show
whether a given scheme captures some of the structure of natural
language text. We then use this method to support a proposal to decompose
the parsing task into computationally more tractable subtasks. This
approach also lends itself to the extraction of predicate/argument
structure.

The principle on which the grammar evaluator is based is derived from
Shannon's original work with letter sequences \cite{shannon}. We show how his 
ideas can be extended to other linguistic entities. We describe a
method of representation that enables the entropy of sentences to be measured 
under different parsing schemes.  
The entropy is a measure, in a certain sense, of the degree of 
unpredictability. If the grammar
captures some of the structure of language, then the relative entropy of
the text should decline after parsing. 
We can thus objectively assess whether parsers that 
accord with some linguistic intuition do indeed capture some regularity
in natural language. 

Natural language can be seen as having a tertiary structure. First, there are 
the relationships between adjacent words, a structure that can be modelled
by Markov processes. Then words can be grouped together into constituents
and these constituents are organized in a secondary structure. Thirdly, there
are relationships between elements of constituents, such as the agreement
between the head of a subject and the main verb. These three levels are 
compatible with levels in the Chomsky hierarchy. 

We need to integrate natural language 
processing at these different levels. 
The work described in this paper uses a method of representation that 
enables primary and secondary structure to be modelled jointly. It concludes 
by suggesting how this approach could facilitate processing at level 
two  and possibly three. 

The paper is organized in the following way. First, we recall Shannon's
original work with letter sequences. Then we describe a method of 
adapting his approach  to word sequences. Next,
we show how this is not an adequate model for natural language sentences,
but can be extended. Using the new representation we can model syntactic
constituents, and parsing a sentence is taken to be finding their location.
Then we show how the entropy of parsed and unparsed sentences is measured.
If the entropy declines after parsing, this indicates that some of the 
structure has been captured.

Finally, we apply this entropy evaluator to show that one particular parsing 
method effectively decomposes declarative sentences into three sections. These
sections can be partially parsed separately, in parallel, thus reducing the 
complexity of the parsing task.

\section{Shannon's work with letter sequences}
Shannon's well known work on characteristics of the English language examined
the entropy of letter sequences.
He produced a series of approximations to the
entropy $H$ of written English, which successively take
more of the statistics of the language into account

$H_0$  represents the average number of bits required to determine a
letter with no statistical information. $H_1$ is calculated with
information on single letter frequencies; $H_2$ uses information on the
probability of 2 letters occurring together;  $H_n$,  called
the n-gram entropy,  measures the amount of entropy with
information extending over $n$ adjacent letters of 
text.\footnote{This notation is derived from that used by Shannon. It differs
from that used by Bell, Cleary and Witten \cite{bell}. }
As $n$ increases from 0 to 3, the n-gram entropy declines: the degree of 
predictability is increased as information from more adjacent letters is taken
into account. If $n -1$ letters are known, $H_n$ is the conditional entropy of 
the next letter, and is defined as follows.
\begin{quote}
$b_i$ is a block of $n-1$ letters, $j$ is an arbitrary letter following $b_i$

$p(b_i,j)$ is the probability of the n-gram $b_i,j$

$p_{b_i}(j)$ is the conditional probability of letter $j$ after block $b_i$,
that is $p(b_i,j) \div p(b_i)$
\end{quote}
\begin{eqnarray*}
H_n & = & - \sum_{i,j} p(b_i,j) \ast log_2 p_{b_i}(j) \\
    & = & - \sum_{i,j} p(b_i,j) \ast log_2 p(b_i,j)  
          + \sum_{i,j} p(b_i,j) \ast log_2 p(b_i) \\
    & = & - \sum_{i,j} p(b_i,j) \ast log_2 p(b_i,j) 
          + \sum_{i} p(b_i) \ast log_2 p(b_i) \\
\end{eqnarray*}

since $\sum_{i,j}p(b_i,j) = \sum_ip(b_i)$

An account of this process can also be found in \cite{cover2}.

Now, the entropy can be reduced if an extra character representing a
space between words is introduced, and the probability of n-grams occurring
is taken into account.  Shannon says ``a word is a 
cohesive group of letters with strong internal statistical influences'' so the
 introduction of the space captures some of the structure of the letter 
sequence.

Let $H'$ represent the entropy measures of the 27 letter alphabet.
By introducing an extra element, the number of choices has increased, so,
without any information on probabilities,
$H'_0 > H_0$. However, if $n > 0$, then $H'_n < H_n$. The space will be more 
common than other characters , so $H'_1 < H_1$. Where $n > 1$ the 
statistical relationships of neighbouring elements are taken into account. 
More of the structure of letter sequences is captured, so entropy declines.
\begin{table}[hbt]
\begin{center}
\begin{tabular}{|lcccc|}
\hline
           & $H_0$ & $H_1$ & $H_2$ & $H_3$ \\
26 letter  & 4.70  & 4.14  & 3.56  & 3.3 \\
27 letter  & 4.76  & 4.03  & 3.32  & 3.1 \\
\hline
\end{tabular}
\caption{ Comparison of entropy for different n-grams, with and without 
representing the space between words}
\end{center}
\end{table}

\section{Representing parsed and unparsed text}

Now, this type of analysis applied to strings of letters can also be
applied to strings of words.
However, in order to make this approach computationally feasible we
need to  partition an indefinitely large vocabulary into a  
limited number of part-of-speech classes. We have to map a large number
of words onto a much smaller number of tags. By taking this step we loose
much information: the process is not reversible. However, we aim to
retain the information that is needed for this particular stage in 
the process. The additional information in the words themselves 
can be held for future reference.

Sometimes, the  allocation of part-of-speech tags has
 been considered a step in parsing. However, we are looking for
syntactic structure and  call the strings of tags the unparsed text. 

Now, at the primary level text can be modelled as
a sequence of tags, and Shannon's type of
analysis can be extended to word sequences. Punctuation marks can also be
mapped onto tags. An experiment with the LOB corpus showed that for 
sequences of  parts-of-speech tags $H_2$ and $H_3$ are usually
 slightly lower if punctuation is included in an enlarged tagset.

 However, there is more structural information to be 
extracted. Our linguistic intuition suggests that there are 
constituents, cohesive groups of  words with internal statistical influences.
The entropy indicator will show objectively whether this intuition is well 
founded.

Furthermore, the statistical patterns of tag sequences can be
disrupted at the boundaries of constituents.  
Consider the probability of part-of-speech tags following
each other: some combinations are ``unlikely'', such as  {\em noun - pronoun} 
and {\em verb -  verb} but they may occur at clause and phrase
boundaries in sentences like
\begin{sf}
\begin{quote}
  The shirt  he wants  is in the wash. \\
\end{quote}
\end{sf}
which maps onto tags
\begin{sf}
\begin{quote}
determiner~~ noun~~ pronoun~~ verb~~ verb~~ preposition~~ determiner~~ noun~~ full stop
\end{quote}
\end{sf}

An important step extends the representation to handle this. The embedded 
clause is delimited by inserting
boundary markers, or hypertags, like virtual punctuation marks. We represent
the sentence as
\begin{sf}
\begin{quote}      
  The shirt \{ he wants \} is in the wash.
\end{quote}
\end{sf}

The pairs and triples generated by this string  would exclude 
{\em noun - pronoun}, {\em noun - pronoun - verb}
but include, for instance,
{\em noun - hypertag1}, {\em noun - hypertag1 - pronoun}.
The part-of-speech tags have probabilistic relationships
with the hypertags in the same way that they do with each other.
We can measure the entropy of the sequence with the opening and closing 
hypertags included.
If their insertion has captured some of the structure
the  bipos and tripos entropy should be reduced.

Each class of syntactic elements has a distinct pair
of hypertags. Applying automated parsers, one type of syntactic element
is found at a time. 
In this particular case of locating an embedded clause, the insertion
of hypertags can be seen as representing ``push'' and ``pop'' commands.
One level of embedding has been replaced.

This approach can be contrasted with the process of text compression. 
In well compressed text the structure should be extracted so that the 
output is ``whitened'', or appears random \cite[chapter 10]{bell}. 
In the process described
here the insertion of virtual markers, the hypertags,  converts segments 
of a sequence
with very weak probabilistic relationships into segments where the 
elements are subject to much stronger probabilistic relationships.

It is interesting to note in passing that we commonly assume that small
children can process embedded structures without difficulty. Books
for young children usually have limited vocabularies and short sentences,
but this type of construction is not deliberately avoided.
Thus, the first page of  ``Jack and the Beanstalk'', published by
Ladybird, has the sentence ``All we have is one cow''.

\section{Entropy measures}
\label{ent measure}

Introducing hypertags can be seen as analogous to adding a space symbol to
the alphabet. Parsing is the process of inserting the hypertags. We then
measure the entropy of the tag sequence with and without the hypertags. If 
the entropy of the parsed text is lower than that of the plain text, as in
the circumstanes described below, then some of the structure of the language 
is captured.

We apply this theory to a corpus of text, taken from
engine maintenance manuals. We propose different structural markers, and
measure the resulting entropy. Note that the absolute entropy levels
depend on a number of variable factors. We are interested in comparative
levels, and thus use the term {\em entropy indicators}. 

There is a relationship between tagset size, distribution of tags, number
of samples and entropy. For instance, as tagset size is decreased entropy
declines, but at the same time grammatical information may be lost. 
We have to balance the requirement for a small tagset against the need to 
represent separately each part-of-speech with distinct syntactic behaviour.  
Another approach to entropy reduction, which would not be
helpful, is to expand one element into several that always, or usually,
occur together. For instance, we can reduce the entropy by mapping every
instance of  {\em determiner} onto {\em hypertag1 determiner hypertag2}.

We use linguistic intuition to propose constituents, substrings of tags with
certain  characteristics that suggest they should be grouped together. Then
we investigate the entropy levels of tagged text for the following cases
\begin{enumerate}
\item No hypertags (suffix p: plain)
\item Arbitrarily placed hypertags: in each sentence before tag  
position 2, after tag  position 5 (suffix a)
\item Hypertags before and after determiners (suffix d) 
\item Hypertags delimiting noun groups (suffix n)
\item Hypertags delimiting subject (suffix s)
\item Hypertags delimiting subject and noun groups (suffix sn)
\end{enumerate}
A noun group is taken to be a noun immediately preceded by an optional number
of modifiers, such as ``mechanical stop lever'' or just ``lever''.  

\subsection*{The subject and embedded clauses}
The boundaries of phrases and clauses often coincide with the boundary
of the subject. As we have the ALPINE system (described below) to 
automatically locate the subject 
we decided to investigate this constituent, rather than embedded phrases
and clauses directly. The output of ALPINE was manually checked for the 
current exercise. We expect that the insertion of hypertags
to demarcate the subject will lower the entropy.

\subsection*{Results}
The data consisted of 351 declarative sentences from manuals from Perkins
Engines Ltd. Average sentence length is 18 words, counting punctuation
marks as words. The tagset had 32 members, including 4 hypertags, so 
$H_0 = 5.0$. The n-grams analysed were pairs and triples.
 Using automated parsers previously
developed, the data was  prepared automatically, but then manually checked.
A summary of results obtained is given in Table \ref{res}.

\begin{table}[hbt]
\begin{center}
\begin{tabular}{|clccc|}
\hline
    & text & $H_1$ & $H_2$ & $H_3$ \\
\hline
1 & text-p & 3.962 & 2.659 & 2.132 \\
\hline
2 & text-a & 4.135 & 2.689 & 2.077 \\
\hline
3 & text-d & 4.086 & 2.123 & 1.722 \\
\hline
4 & text-n & 3.981 & 2.038 & 1.682 \\
\hline
5 & text-s & 4.135 & 2.472 & 1.997 \\
\hline
6 & text-sn & 4.142 & 1.943 & 1.612\\  
\hline
\end{tabular}
\caption{ Entropy measures for text  with different structural markers
\label{res}}
\end{center}
\end{table}

For interest, some text from Shannon's article was also processed in the same
way, and produced results in line with these.

Recall that we are interested in the movement of the entropy measure, and 
do not claim to attach significance to the absolute values. We ask a question 
with a ``yes'' or ``no'' answer: does the entropy decline when the parsing
scheme is applied. However, note the results of 6, which combines schemes 
4 and 5,
that is marking both the noun groups and the subject. We see that the decline
 in entropy $H_2$ and $H_3$ is greater than for either scheme separately. 

We see from Table \ref{res} that the arbitrary placement of
hypertags did indeed increase the entropy. As expected, the placement
either side of the determiner reduced the entropy, but since these
three elements always occur together this is not the result of
capturing language structure. The interesting result comes from
comparing line 1 with lines 5 and 6 of Table \ref{res}. The
placement of the hypertags around the subject, with or without also
locating noun groups, reduces the entropy. The components of the
subject are variable, and in this case the reduction in entropy
indicates that some of the structure has been captured.

\section{Applying these results to decompose the parsing task}

Now that it is technically feasible to locate the subject of a sentence
automatically, we may be able to take advantage of this to reduce the
complexity of parsing English text.
In the corpus used the length of the subject varied from 1 to 12 words,
the length of the pre-subject from 0 to 15 words.
As an example of  subject  location consider these sentences
 from Shannon's paper which would be represented as
\begin{sf}
\begin{quote}
In a previous paper \{ the entropy and redundancy of a language \} have
been defined.

If the  language  is translated into  binary digits  in the  most 
efficient way  , \{ the  entropy  \} is the  average number  of 
 binary digits  required per  letter  of the  original language .
\end{quote}
\end{sf}

\begin{figure}[hbt]
\begin{center}
\epsfig{figure=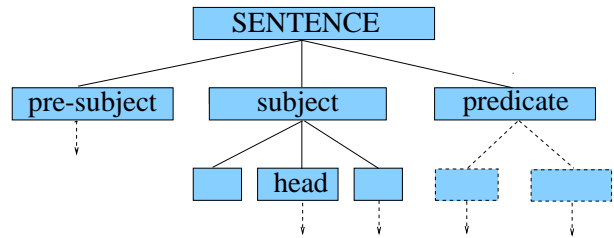,width=8cm}
\end{center}
\caption{\sf Decomposition of the sentence into syntactic constituents 
\label{harchy}}
\end{figure}

Now, locating the subject effectively decomposes a declarative sentence into 
three sections: see Figure \ref{harchy}.
Of course the first 
 section can be empty. Imperative sentences can also be processed in this way,
 the lack of an  explicit subject being represented by an empty subject
section. 

Almost all declarative sentences can be decomposed in this way \cite{lyon14}.
On examining these concatenated sections we note that other constituents 
are contained within them and do not cross the boundaries between them. 
An element or constituent in one section can have 
dependent links to elements in other sections, such as  agreement
between the head of the subject and the main verb.  
However, the constituents themselves - clauses, phrases, noun groups -  
are contained within one section. Therefore, once the three sections 
have been located, they can then be partially processed 
separately, in parallel. The complexity of the parsing task can be reduced
by decomposing a declarative sentence  as a preliminary 
move.

The ALPINE parser that finds the subject, and thus decomposes the sentence,
is being developed. A prototype is available via telnet and readers
are invited to access it and 
try  their own text. For details  contact the authors. ALPINE
is described in \cite{lyon9,lyon11}, and other papers at 
{\tt ftp://www.cs.herts.ac.uk/pub/caroline}. 

\section{Conclusion}

We have shown that entropy indicators can be used to support parsing schemes
based on linguistic intuition. Entropy measures have been used to determine 
the most effective
representation for a formal language \cite{lari}. We suggest that
they can also be used to evaluate representations for natural language.

In particular, the entropy indicator supports the top down
decompostion of a sentence
 into three concatenated segments that can be partially processed separately. 
Since many automatic parsers have difficulty processing longer sentences,
we suggest that this decomposition could facilitate the operation of
other systems. 

Another advantage of this approach to parsing is that it lends itself to
the extraction of predicate/argument structure \cite{www.upenn1}. 
After the subject has
been located the main verb will be found in the predicate, and then the
object or complement. With the head of the subject found, we
then have the raw material from which we can begin to
extract the predicate/argument structure. This  approach is a basis for
beginning to address semantic questions.

\bibliography{bib}
\bibliographystyle{unsrt}
\end{document}